\documentclass[a4paper,12pt]{article}
\usepackage[utf8x]{inputenc}
\usepackage{setspace}
\usepackage{graphicx}
\graphicspath{ {./figs/} }

\usepackage[table,xcdraw]{xcolor}
\usepackage{amsmath} 
\usepackage{amssymb} 
\usepackage{graphicx} 
\usepackage{subcaption}
\usepackage{verbatim}
\usepackage{mathtools}
\usepackage{subcaption}
\usepackage[titlenumbered,ruled,linesnumbered]{algorithm2e}
\usepackage{todonotes}
\usepackage{booktabs}
\usepackage{breakcites}

\usepackage[UKenglish]{babel}

\usepackage{hyperref}
\hypersetup{
    citecolor=black,
    filecolor=black,
    linkcolor=green,
    urlcolor=black
}
\usepackage{subcaption}
\usepackage{flexisym}
\usepackage{listings}
\usepackage{color}

\definecolor{dkgreen}{rgb}{0,0.6,0}
\definecolor{gray}{rgb}{0.5,0.5,0.5}
\definecolor{mauve}{rgb}{0.58,0,0.82}

\begin{document}

\begin{titlepage}

\center 


\vspace{-1cm}
\includegraphics[width=4cm]{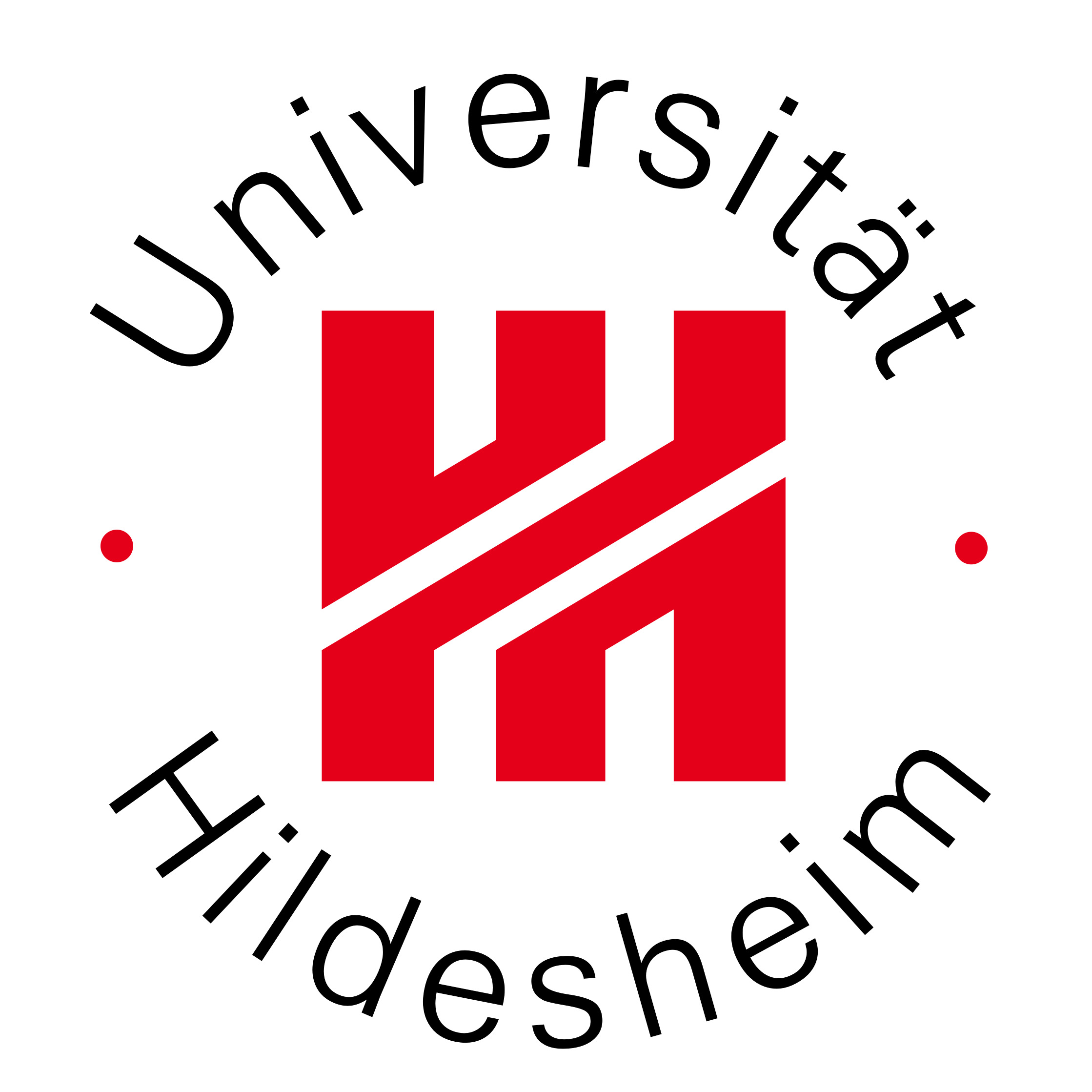}
\vspace{2cm}


{ \Large \bfseries Deep Neural Representation Learning on Dynamic
Graphs via Self-Attention and Convolutional Neural Networks}\\  
\vspace{0.5cm}



\begin{minipage}{0.49\textwidth}
\begin{flushleft} \large
\emph{Authors:}\\
Ahmad  \textsc{Hafez} \\ 
Atulya  \textsc{Praphul} \\ 
Yousef  \textsc{Jaradt} \\ 
Ezani  \textsc{Godwin} \\ 
\end{flushleft}
\end{minipage}
~
\begin{minipage}{0.46\textwidth}
\begin{flushright} \large
\emph{Supervisors:} \\
Ahmed \textsc{Rashed} \\ 

\end{flushright}
\end{minipage}\\
\vspace{1cm}


{ \today}\\ 
\vspace{1cm}

{ \large \bfseries Student Research Project 2020-2021}\\  
\vspace{0.3cm}
\textsc{\Large Master of Science in Data Analytics}\\
\vspace{1cm}
\textsc{\large Wirtschaftsinformatik und Maschinelles Lernen}\\ 
\vspace{0.3cm}
\textsc{\large Stiftung Universität Hildesheim}\\ 
\vspace{0.3cm}
\textsc{\large Universitatsplätz 1, 31141 Hildesheim}\\ 
\vspace{0.3cm}

\vfill 

\end{titlepage}

\setcounter{secnumdepth}{1}

\setcounter{tocdepth}{2}
\newpage


\begin{abstract}

Learning node representations on temporal graphs is a fundamental step to learn real-word dynamic graphs efficiently. Real-world graphs have the nature of continuously evolving over time, such as changing edges weights, removing and adding nodes and appearing and disappearing of edges, while previous graph representation learning methods focused generally on static graphs. We present ConvDySAT as an enhancement of DySAT \cite{sankar2020dysat}, one of the state-of-the-art dynamic methods, by augmenting convolution neural networks with the self-attention mechanism, the employed method in DySAT to express the structural and temporal evolution. We conducted single-step link prediction on a communication network and rating network, Experimental results show significant performance gains for ConvDySAT over various state-of-the-art methods.

\vfill
\end{abstract}

\newpage

\tableofcontents

\newpage

\listoffigures

\listoftables

\pagenumbering{arabic}

\section{Introduction}

Learning embeddings of nodes in graphs is considered vital learning problem due its large applicability to various domains, such as 3D models \cite{fathy2020temporalgat}, knowledge bases \cite{li2017attributed}, bioinformatics \cite{grover2016node2vec} and social media \cite{perozzi2014deepwalk}. The key objective is to learn the most informative low-dimensional representations to capture the structural properties among the nodes in any graph. Representing nodes in a low-dimensional vector form helps to apply graph analysis tasks easily and efficiently, such as clustering \cite{cao2016deep}, graph visualisation \cite{wang2016structural} and link prediction \cite{grover2016node2vec}. Previous graph embedding methods focused for the most part on static graphs \cite{chen2018fastgcn,grover2016node2vec,hamilton2017inductive,kipf2016semi,perozzi2014deepwalk,velivckovic2017graph}, in which it is assumed there is no relation-evolving among the nodes. Nevertheless, most of the real-world graphs are dynamic by nature, where there is a constant evolving over time among the nodes.\\\\
Learning the embedding is challenging owing to the continuous-time evolving nature of the real-world graphs, where nodes can emerge or split every time step and consequently new links can be introduced or removed. Therefore, the embeddings are not only required to capture the properties of the graph, but also to capture temporal evolution over time.\\\\
Since 2017\cite{li2017attributed}, novel methods have been introduced to learn the dynamic graphs. They are usually categorized into two groups: imposing a temporally regularized weights that enforces smoothness of node representations from adjacent time steps \cite{zhou2018dynamic,zhu2016scalable} and employing recurrent neural network \cite{goyal2020capturing,hasanzadeh2019variational} to ease the issue of temporal reasoning with summarizing historical snapshots via hidden states.”  However, both methods partly fail to learn due to the inability of learning when there is large change over every time step and requiring enormous amount of data to perform well, respectively.
Recently, attention mechanism has reached a remarkable achievement in sequential learning tasks \cite{bahdanau2014neural,velivckovic2017graph,yu2018qanet} extended the idea to graphs by enabling the nodes in the graph to attend over the neighbouring nodes, outperforming most of the state-of-the-art static graph embedding methods.
In this paper, our work is inspired mainly by three recent papers \cite{li2019enhancing,sankar2020dysat,velivckovic2017graph} . While \cite{velivckovic2017graph} leveraged self-attention mechanism to learn the structural proximity of the nodes, \cite{sankar2020dysat} stretched the self-attention mechanism to learn the temporal evolution by attending over the historical representation of every node. However, as \cite{li2019enhancing} pointed out, the canonical transformer is agnostic of local context since the query key matching is only point-wise. Therefore, we employed convolutional layers to be more aware of the local context.


 \section{Related Work}

\subsection{Static graph embedding}

SGE methods could be categorized into four groups, matrix factorization-based methods \cite{belkin2001laplacian}, deep autoencoder-based methods \cite{cao2016deep,wang2016structural} and skip-gram-based methods \cite{cao2015grarep,grover2016node2vec,perozzi2014deepwalk,tang2015line} and recently self-attention-based methods \cite{velivckovic2017graph},yet these methods cannot attend to all real-world network situations, such as changing edges weights, removing and adding nodes.

\subsection{Dynamic graph embedding}
Expressing time evolving in dynamic graph embedding defined in two common ways: Snapshot sequence \cite{leskovec2007graph} and Timestamped Graph \cite{trivedi2017know}. The latter express the evolving as a continuous-time function and usually reach better accuracy, Nevertheless, it is computationally expensive. The snapshot sequence is a less-expensive discrete mechanism with taking snapshots every specific amount of time.\\\\

The existent methods mostly base their work on one of the baseline methods to adapt it to work with dynamic environment such as \cite{fathy2020temporalgat,sankar2020dysat,xu2020inductive}. The most recent snapshot-sequence methods \cite{fathy2020temporalgat,sankar2020dysat} based their  work on \cite{velivckovic2017graph} as a static layer. While \cite{sankar2020dysat} extended self-attention mechanism to express the temporal evolution \cite{fathy2020temporalgat} used Temporal Convolutional Networks \cite{bai2018empirical} as a temporal layer. However, some baseline methods such as \cite{perozzi2014deepwalk} could be also considered as dynamic algorithms.

\subsection{Self-attention mechanism}

State-of-the-art NLP work has shown the considerable addition that self-attention mechanism could do in both efficiency and performance. Self-attention was used along two dimensions by \cite{sankar2020dysat,velivckovic2017graph} to learn the node representation. 
\begin{itemize}
    \item Structural neighbourhood: Through self -attentional aggregation, structural attention is able to obtain features from local node neighbourhoods in each snapshot.
    \item Temporal dynamics: temporal attention captures the change that happened in the network through weighting historical representations
 
\end{itemize}

\subsection{Enhancement on self-attention}
One of the major limitations of self-attention with time series data that it is agnostic of the local context \cite{li2019enhancing}. To ease this issue, convolutional layers were used with kernel size more than 1 to transform the outputs of the structural layer (with proper padding) to be more aware of the local context and capture the long-term dependencies.

\subsection{Combining CNNs and RNNs to process long sequences}
\begin{figure}[h]
    \centering
	\includegraphics[scale=0.2]{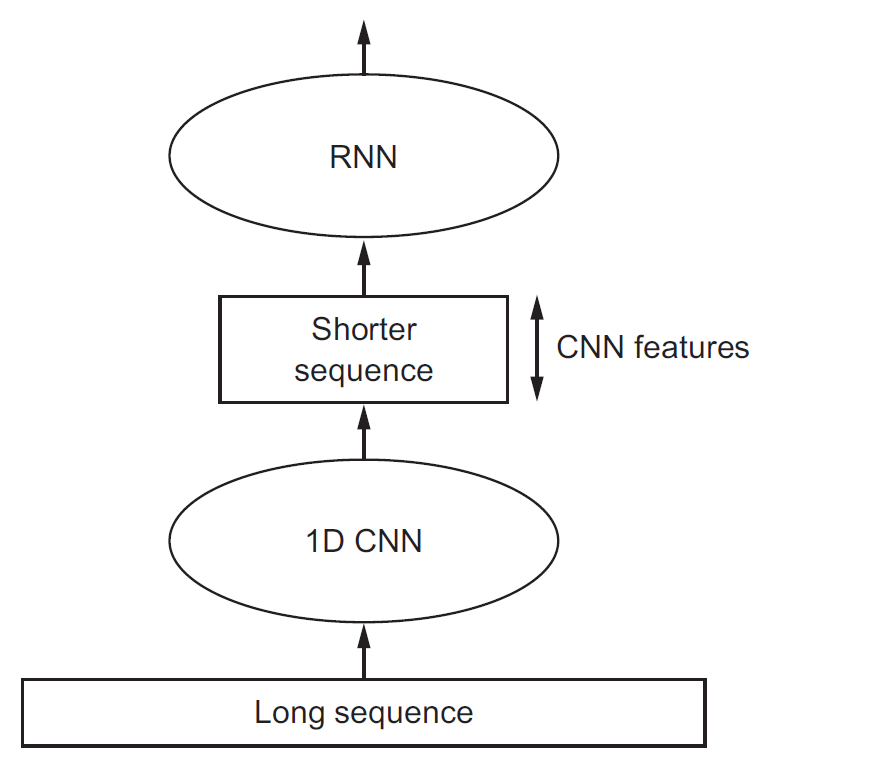}
    \caption{Combining CNNs and RNNs to process long sequences.}
    \label{fig:cnn_rnn}
\end{figure}
 One of the experiments that was carried out is CNN-LSTM temporal layer. The method is not seen often in research papers because it is  not a well-known technique to process long sequences\cite{chollet2018deep}.  CNNs are valuable to time series forecasting and link prediction but  not  aware of the timestamps order. In order to tackle this issue CNNs are combined with the order-sensitive RNNs. The strategy depends on using CNN as a preprocessing step  before  RNN.
 

\section{Problem Definition}
In snapshot-based methods the problem is defined as a sequence of observed static graph snapshots $G_1, G_2, …, G_T$ where $T$ is the number of time steps. A Graph at specific time t (snapshot) is represented by $G_t = (V_t, E_t)$, where $V_t$ and $E_t$ represent the nodes and the links (edges) respectively. The aim is to learn effective latent representation $e_v^t \in R^d$ for each node $v\in Vat$ timesteps $t= {1, 2, …, T}$.

\section{Convolutional DySAT}
In this section, we would be presenting the major components and building blocks of ConvDySAT. ConvDySAT has four modules from top to bottom: The structural attention block; convolutional layer; Temporary self attention; and the graph context prediction. Our key improvement in this project is the addition of the convolutional layers right before the temporal self attention that enables effective extraction of rich features which in turn improved the performance of the model. The structural, convolutional, and temporal layers together model a better graph evolution, and can realize graph neural networks of arbitrary complexity through layer stacking. Our model ConvDySAT was built on these four modules which we present below.

\subsection{Structural Self-Attention}
The data fed into this layer is a graph snapshot $\mathcal{G} \in \mathbb{G}$ and a set of input node representations $x_v \in R^D, \forall v \in \mathcal{V}$ where  $D$ is the input embedding dimension. The input to the initial layer is set as one-hot encoded vectors for each node. The output is a new set of node representations $Z_v \in R^F, \forall v  \in \mathcal{V}$ with $F$ dimensions, that capture the local structural properties in snapshot $\mathcal{G}$.\cite{sankar2020dysat} \\\\
The structural self-attention layer attends over the immediate neighbours of a node (in a snapshot $\mathcal{G}$), by computing attention weights as a function of their input node embeddings. \\

\begin{equation}
\sigma\left(\sum_{u \varepsilon \mathbb{N}} \alpha_{u} W^{S} x_{u}\right), \alpha_{u}=\frac{\exp \left(e_{u}\right)}{\sum_{w \varepsilon \mathbb{N}} \exp \left(e_{w}\right)}
\end{equation}\\

\begin{equation}
e_{u}=\sigma\left(A_{u} \cdot a^{T}\left[W^{S} x_{u}|| W^{S} x\right]\right) \quad \forall(u,) \varepsilon \varepsilon
\end{equation}\\

Where $\mathbb{N}_v= u \in \mathcal{V} \colon (u,v)$ is the set of immediate neighbours of node $v$ in snapshot $\mathcal{G} ; W^s \in R^{F\times{D}}$ is a shared weight transformation applied to each node in the graph; $a \in R^2D$ is a weight vector parametrizing the attention function implemented as feed-forward layer; $||$ is the concatenation operation and $\sigma(\centerdot)$ is a non-linear activation function. $A_{uv} $ is the weight of the link $(u,v)$ in the current snapshot $\alpha_{uv} $. $uv$ are learned coefficients obtained by the contribution of each node in $\mathcal{V}$ which indicate the contribution of node $u$ to node $v$ \cite{sankar2020dysat}.
\subsection{Convolutional Layer}
The convolutional layer extracts vital information from the inputs gotten from the structural attention layer. By employing casual convolutions, we are able to produce feature rich queries and keys to be passed to the masked self attention for fine-grained node representation learning. Query key matching aware of local context, e.g. shapes, can help the model achieve lower training loss and further improve the model's forecasting accuracy.\\
\begin{equation}
K_{i j m}, Q_{i j m}=\sum_{K=0}^{K-1} \sum_{P=0}^{H-1} \sum_{q=0}^{H-1}{ }_{i+p, j+q, k^{h} p q k m}^{(l-1)}+b_{i j m}
\end{equation}
For k =2\\

\begin{equation}
V_{i j m}=\sum_{K=0}^{K-1} \sum_{P=0}^{H-1} \sum_{q=0}^{H-1}{ }_{i+p, j+q, k^{h} p q k m}^{(l-1)}+b_{i j m}
\end{equation}
For k =1\\
With the query-key from the above convolutions performed on the node representations from  the structural attention layer we are able to effectively capture the structure of every respective node in the graph.

\subsection{Masked Attention(Temporal Self-Attention Layer)}
The masked attention layer captures the temporary evolution of the dynamic graph. The input to this layer is a sequence of representations for a particular node learned by the convolutional layer at different time steps. The input representations from the convolutional layer are assumed to sufficiently capture local structural information at each time step, which enables a modular separation of structural and temporal modelling. \\

For each node $v$, we define the inputs as ${x_v^1,x_v^2,x_v^3, \dotsb, x_v^T} , x_v^t \in  D^\prime$ where $T$ is the total number of time steps and $D^\prime$ is the dimensionality of the input representations. The output of this layer is the new representation for $v$ at each time step, i.e, $Z_v = {Z_v^1,Z_v^2,\dotsb,  Z_v^T}, Z_v^t \in  RF^\prime$ with dimensionality $F^\prime$.
The main aim of the masked multi-head attention is to capture the temporal variations in the graph structure over multiple time steps. The input representation of node $v$ at time-step $t$ , $x_v^t$ , encodes the current local structure around $v$. We use $x_v^t$ as the query to attend over its historical representations $(< t )$, tracing the evolution of the local neighbourhood around $v$.\cite{sankar2020dysat}\\\\

The main aim of the masked multi-head attention is to capture the temporal variations in the graph structure over multiple time steps. The input representation of node $v$ at time-step $t$ , $x_v^t$ , encodes the current local structure around $v$. We use $x_v^t$ as the query to attend over its historical representations $(< t )$), tracing the evolution of the local neighbourhood around $v$.\\
\begin{equation}
\beta(X W), \quad \beta^{i j}=\frac{\exp \left(e^{i j}\right)}{\sum_{w \in \mathbb{N}}^{T} \exp \left(e^{i j}\right)}
\end{equation}\\

\begin{equation}
e^{i j}=\left(\frac{\left(\left(X W_{q}\right)\left(X W_{k}\right)^{T}\right) i j}{\sqrt{F^{\prime}}}+M_{i j}\right)
\end{equation}\\

Where $\beta_v \in R^{T\times{T}}$ is the attention weight matrix obtained by the multiplicative attention function and $M \in R^{T\times{T}}$ is a mask matrix with each entry $M_{ij} \in {−\infty, 0}$ to enforce the auto-regressive property. To encode the temporal order, we define $M$ as:

\begin{equation}
M_{i j}=\left\{\begin{array}{ll}
0, & i \leq j \\
-\infty, & \text { otherwise }
\end{array}\right.
\end{equation}

\subsection{Multi-Faceted Graph Evolution}
Our approach can sufficiently capture a single type or facet of graph evolution by stacking structural,convolutional and masked attention layers. However, real-world dynamic graphs typically evolve along multiple latent facets, e.g., evolution of users movie-watching preferences across different genres (such as sci-fi, comedy, etc.) exhibit significantly distinct temporal trends. Thus, we endow our model with expressively to capture dynamic graph evolution from different latent perspectives through convolutional self multi-head attentions. Multi-head attention, which creates multiple independent instances of the attentional function named attention heads that operate on different portions of the input embedding, is widely utilized to improve the diversity of attention mechanisms inspired from \cite{sankar2020dysat}.
	
	\subsubsection{Structural multi-head attention}
	Multiple attention heads are computed in each structural attention layer (one per facet), followed by concatenation to compute output representations. As taken from \cite{sankar2020dysat} \\
	\begin{equation}
h=\operatorname{Concat}\left(z_{v}^{1},z_{v}^{2}, \ldots,z_{v}^{H_{S}}\right) \forall \varepsilon \mathcal{V}
\end{equation}\\
	\subsubsection{Multi-head convolutional layers}
	Similar to the structural attention settings, multiple layers of these convolutions are stacked on top each other to produce corresponding query-keys to the masked multihead attention layer. There are no concatenations done at this layer. It produces respective query-keys for every node in the graph to better capture structural patterns \cite{sankar2020dysat}.
	\subsubsection{Masked multi-head self-attention}
	Similar to the structural attention settings, multiple temporal attention heads (or facets) are computed over historical time steps, to compute final node representations.\\
	\begin{equation}
H=\text { Concat }\left(Z_{v}^{1},Z_{v}^{2}, \ldots, Z_{v}^{H_{T}}\right) \forall \varepsilon \mathcal{V}
\end{equation}

where $H_T$ is the number of temporal attention heads, and $H_v \in R^{T\times{F^\prime}} $ is the output of temporal multi-head attentions.

\subsection{ConvDySAT Architecture}
Here we present our neural architecture ConvDySAT for dynamic graph learning and node prediction, that uses the defined structural, convolutional, and masked attention layers as fundamental modules. The input is a collection of T graph snapshots, and the outputs are node representations at each time step. As illustrated in Figure 1, ConvDySAT consists of a structural block which contains the structural attentions followed by convolution layers and mask attention layers which is contained in the temporal block, where each block contains multiple stacked layers of the corresponding layer type. The structural block extracts features from higher-order local neighbourhoods of each node through a self-attentional aggregation and stacking, to compute intermediate node representations for each snapshot. This sequence of node representations then feeds as input to the temporal block, which attends over multiple historical time steps, capturing temporal variations in the graph structure. The outputs of the temporal block comprise the set of final dynamic node representations, which are optimized to preserve the local graph context in each time step.\\

\begin{figure}[h]
    \centering
	\includegraphics[width =  \textwidth]{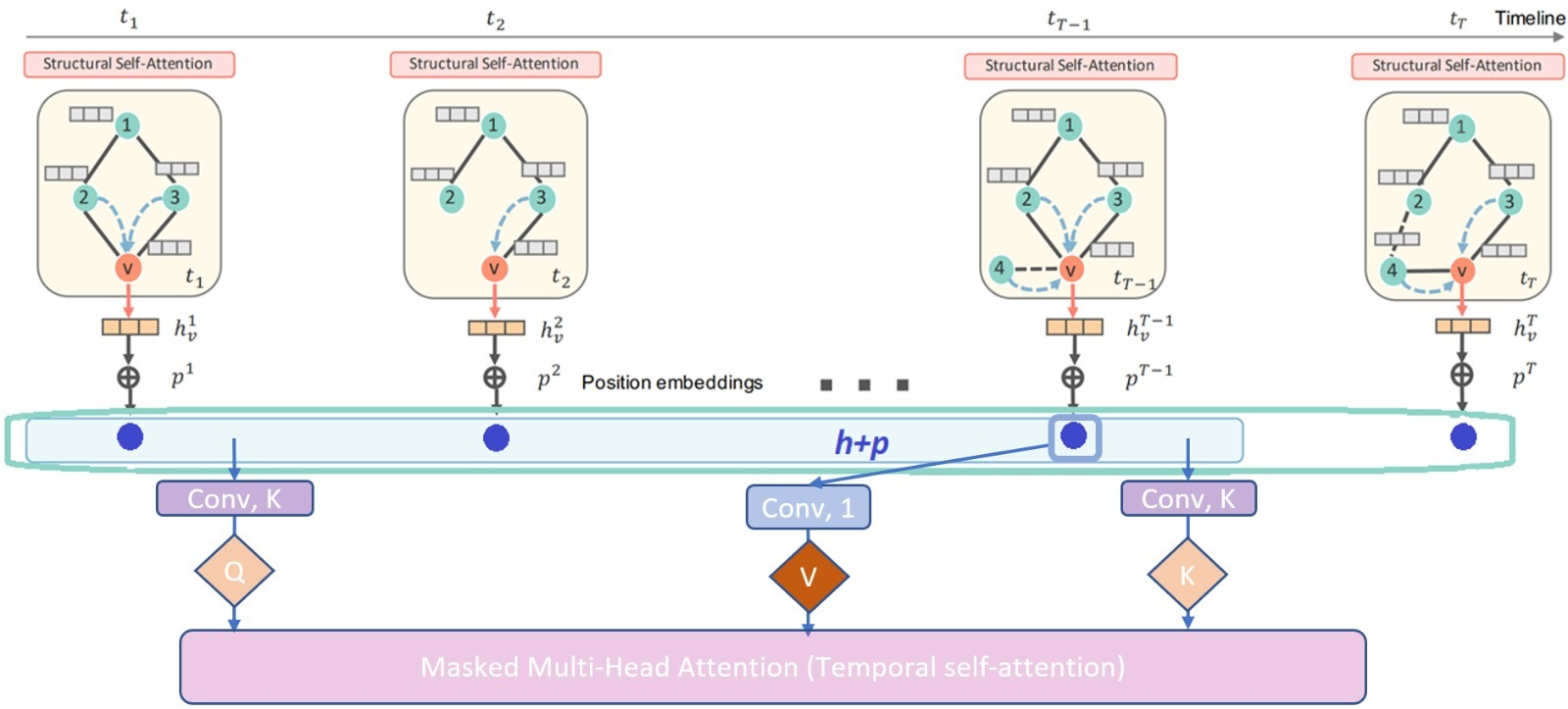}
    \caption{ConvDySAT Architecture.}
    \label{fig:cd_architecture}
\end{figure}

	\subsubsection{Structural attention block}
	To extract features from nodes at different distances just like the baseline we use a block of structural self-attention layer. Here each layer is applied on each graph snapshot with shared parameters, to capture the local structure around a node at each time step. The output of the structural block , $h_v^T$ is then passed to the temporal block.
	
	\subsubsection{Temporal attention block}
	The temporal block contains the position embedding layer, convolutional layer and the masked attention. First, we capture the ordering information in the temporal attention module by using position embeddings $P^1,P^2,P^3, \dotsb ,P^T ,P^t \in R^F$, which embed the absolute temporal position of each snapshot.The position embeddings are then combined with the output of the structural attention block to obtain a sequence of input representations: $h_v^1 +P^1,h_v^2 +P^2,h_v^3 +P^3, \dotsb  ,+h_v^T +P^T$ for node $v$ across multiple time steps. This combination is then passed into the convolutional layers to obtain the respective query-key $(K,Q,V)$ of the input representation. The query-keys from the convolutional layers are then passed into the masked attention layers to produce the final output $e_v^1,e_v^2,e_v^3,\dotsb, e_v^T \forall v \in V $ of the temporal attention block.

	\subsubsection{Graph context prediction}
	In order to capture structural evolution, our objective function preserves the local structure around a node across multiple time steps. We use the dynamic representation of a node $v$ at time step $t$ , $e_u^i$ to preserve local proximity around $v$ at $t$ like the baseline \cite{sankar2020dysat}.\\
	
\begin{equation}
\footnotesize{
\begin{aligned}
L=\sum_{t=1}^{T} \sum_{v \in \mathcal{V}}\left(\sum_{u \in \mathcal{N}_{\text {walk }}^{t}(v)}-\log \left(\sigma\left(<\boldsymbol{e}_{u}^{t}, \boldsymbol{e}_{v}^{t}>\right)\right)\right.
&\left. -w_{n} \cdot \sum_{u^{\prime} \in P_{n}^{t}(v)} \log \left(1-\sigma\left(<\boldsymbol{e}_{u}^{t}, \boldsymbol{e}_{v}^{t}>\right)\right)\right)
\end{aligned}
}
\end{equation}

As you can see in the above equation $N_{walk}^t(v)$ is the set of nodes that co-occur with $v$ on fixed-length random walks at snapshot t , $P_n^t$ is a negative sampling distribution for snapshot $\mathcal{G}^t$, $\sigma$ is the sigmoid function, $< \cdot >$ denotes the inner product operation,and $w_n$ the negative sampling ratio, which balances the positive and negative samples and this is a tunable hyperparameter.
\subsection{CNNLSTM}
We also experimented on some other ideas which proved promising but did not perform as well as ConvDysat. This experiment involves the same process like \cite{sankar2020dysat} but in this case the temporary attention layer was completely replaced by CNNLSTM Layer. Instead of using a temporary masked attention layer, we used CNNLSTM layers which were also stacked up according to the number of time steps. The architecture of CNNLSTM Dysat, involves three main modules just like \cite{sankar2020dysat} . The structural attention layer, the CNNLSTM layer and the graph context prediction. The structural attention layer and the graph context prediction layerare completely the same as the above explained. The only difference is the CNNLSTM.

\begin{figure}[h]
    \centering
	\includegraphics[width =  \textwidth]{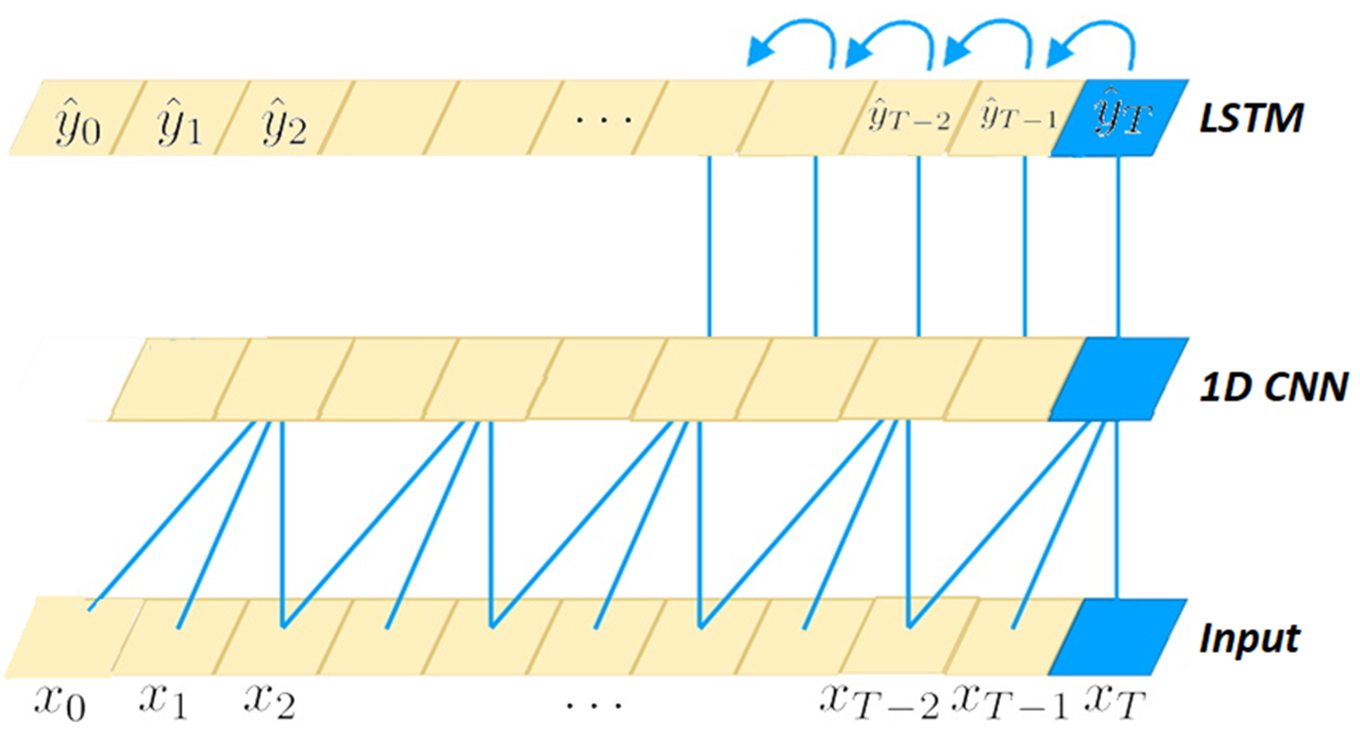}
    \caption{CNNLSTM structure.}
    \label{fig:cd_CNNLSTM}
\end{figure}

CNNLSTM Layer: This layer receives graphical representation as input from the structural attention layer from each individual time step.
This input is passend first into multiple CNN layers to extract feature rich representations and then passes the output to a single layer of LSTM to enforce autoregresiveness, before passing it to the graph context predictor for predictions. This proves to work because with CNN, we were able to Extract feature rich representations, maintain the structural position of each node and their relationship to one another while the LSTM helps to enforce the auto regressive property of the model which made it possible for the model to recognize if a link appeared in a previous time step and the possibility of the link being present in the current timestamp. This Algorithm performs almost as good as ConvDySAT and is better than \cite{sankar2020dysat} in node predictions.

\section{Experiments}
Our experiment consist of step by step comparison of Conv-Dysat to the baseline. We made this comparison successful by using the same budget for both the models. Table \ref{tab:budgets_table} explain the budget we used for each dataset along with the attributes.\\
\begin{table}[h]
\centering
\begin{tabular}{|l|l|l|l|}
\hline
Attribute          & ENRON & YELP   & ML\_10M \\ \hline
Epochs / time step & 200   & 200    & 100     \\ \hline
No. of nodes       & 143   & 6,569  & 20,537  \\ \hline
No. of links       & 2,347 & 95,361 & 43,760  \\ \hline
No. of Time steps  & 16    & 12     & 6       \\ \hline
\end{tabular}
\caption{Experiment Attributes and Budgets}
\label{tab:budgets_table}
\end{table}
\subsection{Datasets}

We have tested against three dynamic graphs: two rating networks of variable sizes and one communication shown in (Table \ref{tab:budgets_table}).
Rating networks considered were Yelp and MovieLens (ML-10M). Yelp consists of links between users and businesses based on the ratings.MovieLens (ML-10M) is formed of user-tag interactions where links connect users with movies tags.
The communication network examined was Enron where links represent email interactions between core employees.
Table \ref{tab:budgets_table} also describe the number of nodes and links present in each of these datasets.

\subsection{Baseline}
We compared against the baseline  which was already compared against other static and dynamic graph embedding methods as follows:
\begin{itemize}
    \item 
node2vec : A static embedding technique uses a second order random walk sampling to learn graph representations.

    \item 
GraphSAGE : A static embedding method of inductive node representation learning framework. Various aggregators like GCN, LTSM, maxpool and meanpool used to reveal the best per each dataset.

	\item

G-SAGE + GAT : A static embedding technique GraphSAGE and the aggregation function: Graph Attentional layer.

\item
GCN-AE : A static embedding method where GCN trained as an autoencoder for link prediction.

\item
GAT-AE : A static embedding technique used to predict link like GAT autoencoder 

\item
DynamicTriad :
A dynamic graph embedding method works on discrete snapshots, it merges triadic closure and temporal smoothness.

\item
DynGEM :
A dynamic graph embedding method works on discrete snapshots, it is a deep neural embedding technique which gradually learn graph autoencoders of different layer sizes.

\item
DynAERNN :
A dynamic graph embedding method works on discrete snapshots that is a deep neural network consists of recurrent and dense layers to capture temporal node evolution

\item
DySAT :
Dynamic network employs GAT as a static layer and self-attention mechanism to capture the temporal graph evolution 
\item
TemporalGAT :
Dynamic network employs GAT as a static layer and temporal convolutional network (TCN) to capture the temporal graph evolution.
 
\end{itemize}

A $20\%$ links were used as a validation set to adjust the hyperparameters of the model.
$25\%$ examples were sampled for training the rest $75\%$ are used for testing and results are averaged across 5 randomized runs to be reported with standard deviation fro both the baseline and our model ConvDySAT. A downstream classifier was trained using Hadamard Operator to calculate a feature vector for a pair of nodes, as suggested by \cite{grover2016node2vec}.
DySAT and ConvDySAT were implemented using Tensorflow and used mini-batch gradient descent with Adam.

\subsection{Result}
The accuracy is computed as the micro and macro AUC score. AUC, that is, the  area under the ROC curve measures the performance  of a  classification problem at various thresholds settings.
Micro AUC is calculated across the link instances from all the time steps. That is, from time step 1 to t.\\
Macro AUC value is the sum of all the AUC value at each time step averaged over the total number of time steps. Thus, it  indicates how good the performance of  a classifier is at each individual time step. The following plots have been made after doing five randomized runs for both the Baseline and Convolutional DySat for 3 different dataset Enron, yelp and ML-10M.

\subsubsection{Enron}

\begin{figure}[h]
    \centering
	\includegraphics[scale=0.4]{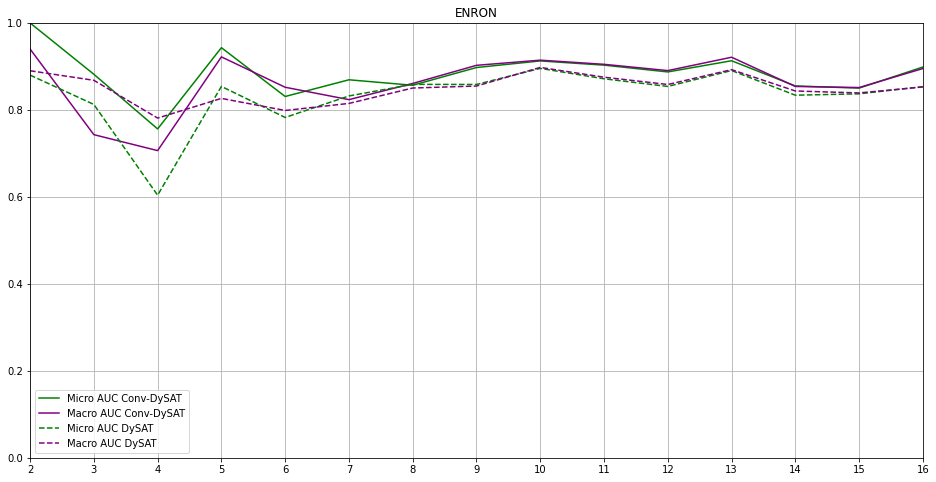}
    \caption{Solid lines represent ConvDySat and dotted line represent the baseline. And their Micro and Macro AUC score respectively.}
    \label{fig:enron}
\end{figure}

This is the result after running both the models on Enron dataset for 200 epochs and 16 time steps. We see around $5\%$ raise in the Micro and Macro AUC value than the baseline. At the last time step(16th) Micro AUC of Convolutional DySat is $89.8\%$ whereas for the baseline is $85.3\%$ and the Macro AUC is $85.2\%$ for baseline and $89.5\%$ for Convolutional DySat.

\subsubsection{Yelp}
	 Conv DySat was run on Yelp dataset for 12 time steps with 200 epochs per time step same as the Baseline. We observe $4-5\%4$ raise in the accuracy. Conv DySat results are reported for two different kernel sizes for the convolutional layer.Test accuracy at $12^{th}$ time step for Kernel size 2, Micro AUC and Macro AUC is $73.8\%$ and $73.7\%$ respectively and for Kernel size 3 Micro AUC is $76.1\%$ and Macro AUC : $76.5\%$

\begin{figure}[h]
    \centering
	\includegraphics[scale=0.4]{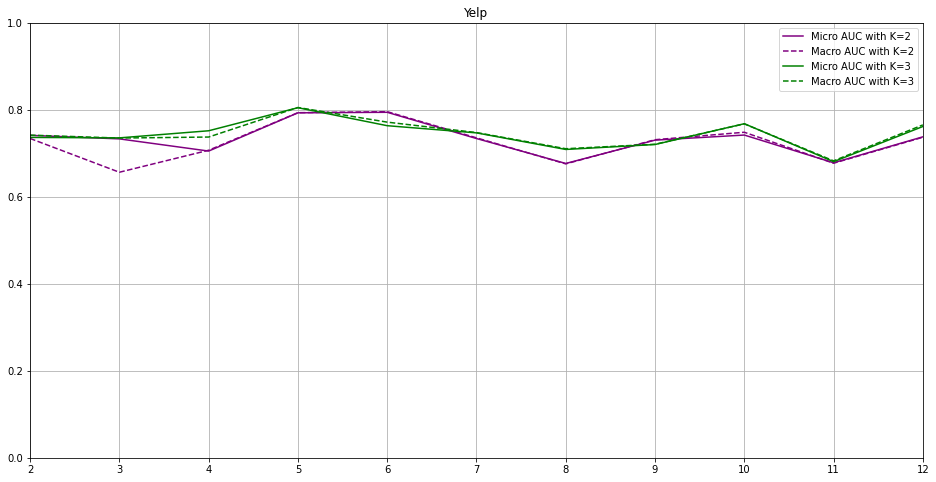}
    \caption{Purple lines represent ConvDySat with kernel 2 and green lines represent ConvDySat with kernel 3. And their Micro and Macro AUC score respectively.}
    \label{fig:yelp}
\end{figure}

	\subsubsection{ML-10M}
	
	Due to the computational expenses the experiment was scaled down to 100 epochs per time step and in total 6 time steps for this dataset. The results are gather for the baseline and proposed model for the same scale. Test accuracy at $6^th$ time step for the baseline is $86.2\%$ of Micro AUC and $75\%$ of Macro AUC and for Conv DySAT Micro AUC received is $88\%$ with $82\%$ Macro AUC.
	
\begin{figure}[h]
    \centering
	\includegraphics[scale=0.4]{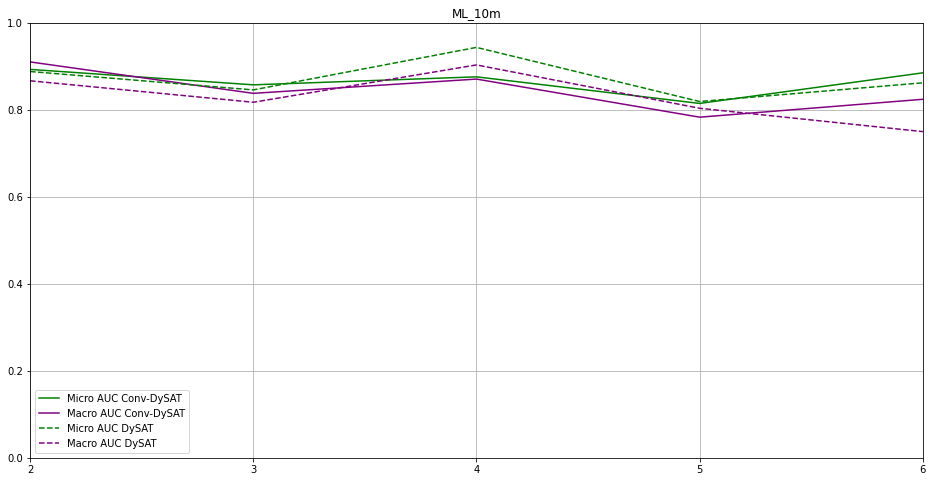}
    \caption{ML-10M trained for 6 time steps on DySAT and ConvDySAT}
    \label{fig:ml}
\end{figure}
\newpage  	
	\subsubsection{Summary of results}
Table \ref{tab:summary_table_1} is a summary of results from various static and dynamic graph embedding method and it shows how ConvDySAT outperforms all these methods.As for Dataset-ML-10M the comparison is made only with the baseline that is the DySAT and the results are as show in Table \ref{tab:summary_table_2} .\\

TemporalGAT in \cite{fathy2020temporalgat} is an attention based Dynamic graph representation learning model that learns the low dimensional feature representation of the graph structure and the temporal changes. This architecture outperforms our chosen baseline DySAT and our proposed method outperforms the TemporalGAT is the following cases: Dataset-Enron accuracy metric-MicroAUC, Dataset-Yelp accuracy metric-MicroAUC and MacroAUC. \cite{zhou2018graph}\\

\begin{table}[h]
\centering
\begin{tabular}{|c|c|c|c|c|}
\hline
\textbf{Method}    & \multicolumn{2}{c|}{\textbf{Enron}}     & \multicolumn{2}{c|}{\textbf{Yelp}}      \\ \hline
\textbf{}          & \textbf{Micro-AUC} & \textbf{Macro-AUC} & \textbf{Micro-AUC} & \textbf{Macro-AUC} \\ \hline
node2vec     & 83.72 ± 0.7 & 83.05 ± 1.2       & 67.86 ± 0.2 & 65.34 ± 0.2 \\ \hline
G-SAGE       & 82.48 ± 0.6 & 81.88 ± 0.5       & 60.95± 0.1  & 58.56± 0.2  \\ \hline
G-SAGE+ GAT  & 72.52 ± 0.4 & 73.34 ± 0.6       & 66.15 ± 0.1 & 65.09 ± 0.2 \\ \hline
GCN-AE       & 81.55 ± 1.5 & 81.71 ± 1.5       & 66.71 ± 0.2 & 65.82 ± 0.2 \\ \hline
GAT-AE       & 75.71 ± 1.1 & 75.97 ± 1.4       & 65.92 ± 0.1 & 65.37 ± 0.1 \\ \hline
DynamicTriad & 80.26 ± 0.8 & 78.98 ± 0.9       & 63.53 ± 0.3 & 62.69 ± 0.3 \\ \hline
DynGEM       & 67.83 ± 0.6 & 69.72 ± 1.3       & 66.02 ± 0.2 & 65.94 ± 0.2 \\ \hline
DynAERNN     & 72.02 ± 0.7 & 72.01 ± 0.7       & 69.54 ± 0.2 & 68.91 ± 0.2 \\ \hline
DySAT        & 85.31       & 86.06             & 70.15 ± 0.1 & 69.87 ± 0.1 \\ \hline
TemporalGAT  & 86.4±0.4    & \textbf{86.8±0.3} & 71.9±0.3    & 70.3±0.2    \\ \hline
\rowcolor[HTML]{93C47D} 
\textbf{ConvDySAT} & \textbf{88.37}     & 86.33              & \textbf{74.37}     & \textbf{74.46}     \\ \hline
\end{tabular}
\caption{Summary of results of Enron and Yelp.}
\label{tab:summary_table_1}
\end{table}

\begin{table}[ht]
\centering
\begin{tabular}{|c|c|c|}
\hline
\textbf{Method}    & \multicolumn{2}{c|}{\textbf{ML=10M}}    \\ \hline
\textbf{}          & \textbf{Micro-AUC} & \textbf{Macro-AUC} \\ \hline
DySAT              & \textbf{87.19}     & 84.79              \\ \hline
\rowcolor[HTML]{93C47D} 
\textbf{ConvDySAT} & \textbf{86.54}     & \textbf{85.06}     \\ \hline
\end{tabular}
\caption{Summary of results of ML-10M.}
\label{tab:summary_table_2}
\end{table}

\section{Discussion}
The aim of this paper is to achieve higher accuracy in Graph embedding for Graph networks that change over time. After deployment of various models like LSTM, CNNLSTM and Transformers we arrived at our baseline paper DySAT. The baseline stacks up the temporal attention layer on top of the structural attention layer to capture the structural and temporal changes in the graph. The ConvDySAT architecture proposes the use of Convolutional layers due to its pattern detection property. ConvDySAT thus outperforms the baseline and all the existing graph embedding methods without much additional cost. We observe a constant increase in accuracy on three different datasets, ENRON, Yelp and ML-10M. The comparison between DySAT and ConvDySAT is done by training both the models on the same datasets with the exact same training budget.

\section{Conclusion}
 In this report we describe a dynamic graph embedding method named ConvDySAT which comprises Temporal attention layer, CNN and structural self attention layer to capture the structure, patterns and dynamic changes in a graph to produce node embeddings with higher accuracy. We implement our model on 3 real world datasets. Our model which is inspired by DySAT, our baseline outperforms it. This indicates how the structural and temporal evolutions of a Dynamic graph is successfully captured better than the baseline and  many other state-of-the-art methods. \\\\
There are certain feasible future expansion of our work. For instance working on a much larger dataset UCI, given the computational cost is affordable. Secondly our model can be expanded for multi-layer dynamic graph networks and multifeatured graph networks like Co-Authorship dataset.

\bibliographystyle{apalike}
\bibliography{./sections/bibliography}

\begin{thebibliography}{}

\bibitem[Bahdanau et~al., 2014]{bahdanau2014neural}
Bahdanau, D., Cho, K., and Bengio, Y. (2014).
\newblock Neural machine translation by jointly learning to align and
  translate.
\newblock {\em arXiv preprint arXiv:1409.0473}.

\bibitem[Bai et~al., 2018]{bai2018empirical}
Bai, S., Kolter, J.~Z., and Koltun, V. (2018).
\newblock An empirical evaluation of generic convolutional and recurrent
  networks for sequence modeling.
\newblock {\em arXiv preprint arXiv:1803.01271}.

\bibitem[Belkin and Niyogi, 2001]{belkin2001laplacian}
Belkin, M. and Niyogi, P. (2001).
\newblock Laplacian eigenmaps and spectral techniques for embedding and
  clustering.
\newblock In {\em Nips}, volume~14, pages 585--591.

\bibitem[Cao et~al., 2015]{cao2015grarep}
Cao, S., Lu, W., and Xu, Q. (2015).
\newblock Grarep: Learning graph representations with global structural
  information.
\newblock In {\em Proceedings of the 24th ACM international on conference on
  information and knowledge management}, pages 891--900.

\bibitem[Cao et~al., 2016]{cao2016deep}
Cao, S., Lu, W., and Xu, Q. (2016).
\newblock Deep neural networks for learning graph representations.
\newblock In {\em Proceedings of the AAAI Conference on Artificial
  Intelligence}, volume~30.

\bibitem[Chen et~al., 2018]{chen2018fastgcn}
Chen, J., Ma, T., and Xiao, C. (2018).
\newblock Fastgcn: fast learning with graph convolutional networks via
  importance sampling.
\newblock {\em arXiv preprint arXiv:1801.10247}.

\bibitem[Chollet et~al., 2018]{chollet2018deep}
Chollet, F. et~al. (2018).
\newblock {\em Deep learning with Python}, volume 361.
\newblock Manning New York.

\bibitem[Fathy and Li, 2020]{fathy2020temporalgat}
Fathy, A. and Li, K. (2020).
\newblock Temporalgat: Attention-based dynamic graph representation learning.
\newblock In {\em Pacific-Asia Conference on Knowledge Discovery and Data
  Mining}, pages 413--423. Springer.

\bibitem[Goyal et~al., 2020]{goyal2020capturing}
Goyal, P., Chhetri, S.~R., and Canedo, A.~M. (2020).
\newblock Capturing network dynamics using dynamic graph representation
  learning.
\newblock US Patent App. 16/550,771.

\bibitem[Grover and Leskovec, 2016]{grover2016node2vec}
Grover, A. and Leskovec, J. (2016).
\newblock node2vec: Scalable feature learning for networks.
\newblock In {\em Proceedings of the 22nd ACM SIGKDD international conference
  on Knowledge discovery and data mining}, pages 855--864.

\bibitem[Hamilton et~al., 2017]{hamilton2017inductive}
Hamilton, W.~L., Ying, R., and Leskovec, J. (2017).
\newblock Inductive representation learning on large graphs.
\newblock {\em arXiv preprint arXiv:1706.02216}.

\bibitem[Hasanzadeh et~al., 2019]{hasanzadeh2019variational}
Hasanzadeh, A., Hajiramezanali, E., Narayanan, K., Duffield, N., Zhou, M., and
  Qian, X. (2019).
\newblock Variational graph recurrent neural networks.
\newblock {\em Advances in neural information processing systems}, 32.

\bibitem[Kipf and Welling, 2016]{kipf2016semi}
Kipf, T.~N. and Welling, M. (2016).
\newblock Semi-supervised classification with graph convolutional networks.
\newblock {\em arXiv preprint arXiv:1609.02907}.

\bibitem[Leskovec et~al., 2007]{leskovec2007graph}
Leskovec, J., Kleinberg, J., and Faloutsos, C. (2007).
\newblock Graph evolution: Densification and shrinking diameters.
\newblock {\em ACM transactions on Knowledge Discovery from Data (TKDD)},
  1(1):2--es.

\bibitem[Li et~al., 2017]{li2017attributed}
Li, J., Dani, H., Hu, X., Tang, J., Chang, Y., and Liu, H. (2017).
\newblock Attributed network embedding for learning in a dynamic environment.
\newblock In {\em Proceedings of the 2017 ACM on Conference on Information and
  Knowledge Management}, pages 387--396.

\bibitem[Li et~al., 2019]{li2019enhancing}
Li, S., Jin, X., Xuan, Y., Zhou, X., Chen, W., Wang, Y.-X., and Yan, X. (2019).
\newblock Enhancing the locality and breaking the memory bottleneck of
  transformer on time series forecasting.
\newblock {\em arXiv preprint arXiv:1907.00235}.

\bibitem[Perozzi et~al., 2014]{perozzi2014deepwalk}
Perozzi, B., Al-Rfou, R., and Skiena, S. (2014).
\newblock Deepwalk: Online learning of social representations.
\newblock In {\em Proceedings of the 20th ACM SIGKDD international conference
  on Knowledge discovery and data mining}, pages 701--710.

\bibitem[Sankar et~al., 2020]{sankar2020dysat}
Sankar, A., Wu, Y., Gou, L., Zhang, W., and Yang, H. (2020).
\newblock Dysat: Deep neural representation learning on dynamic graphs via
  self-attention networks.
\newblock In {\em Proceedings of the 13th International Conference on Web
  Search and Data Mining}, pages 519--527.

\bibitem[Tang et~al., 2015]{tang2015line}
Tang, J., Qu, M., Wang, M., Zhang, M., Yan, J., and Mei, Q. (2015).
\newblock Line: Large-scale information network embedding.
\newblock In {\em Proceedings of the 24th international conference on world
  wide web}, pages 1067--1077.

\bibitem[Trivedi et~al., 2017]{trivedi2017know}
Trivedi, R., Dai, H., Wang, Y., and Song, L. (2017).
\newblock Know-evolve: Deep temporal reasoning for dynamic knowledge graphs.
\newblock In {\em International Conference on Machine Learning}, pages
  3462--3471. PMLR.

\bibitem[Veli{\v{c}}kovi{\'c} et~al., 2017]{velivckovic2017graph}
Veli{\v{c}}kovi{\'c}, P., Cucurull, G., Casanova, A., Romero, A., Lio, P., and
  Bengio, Y. (2017).
\newblock Graph attention networks.
\newblock {\em arXiv preprint arXiv:1710.10903}.

\bibitem[Wang et~al., 2016]{wang2016structural}
Wang, D., Cui, P., and Zhu, W. (2016).
\newblock Structural deep network embedding.
\newblock In {\em Proceedings of the 22nd ACM SIGKDD international conference
  on Knowledge discovery and data mining}, pages 1225--1234.

\bibitem[Xu et~al., 2020]{xu2020inductive}
Xu, D., Ruan, C., Korpeoglu, E., Kumar, S., and Achan, K. (2020).
\newblock Inductive representation learning on temporal graphs.
\newblock {\em arXiv preprint arXiv:2002.07962}.

\bibitem[Yu et~al., 2018]{yu2018qanet}
Yu, A.~W., Dohan, D., Luong, M.-T., Zhao, R., Chen, K., Norouzi, M., and Le,
  Q.~V. (2018).
\newblock Qanet: Combining local convolution with global self-attention for
  reading comprehension.
\newblock {\em arXiv preprint arXiv:1804.09541}.

\bibitem[Zhou et~al., 2018a]{zhou2018graph}
Zhou, J., Cui, G., Zhang, Z., Yang, C., Liu, Z., Wang, L., Li, C., and Sun, M.
  (2018a).
\newblock Graph neural networks: A review of methods and applications.
\newblock {\em arXiv preprint arXiv:1812.08434}.

\bibitem[Zhou et~al., 2018b]{zhou2018dynamic}
Zhou, L., Yang, Y., Ren, X., Wu, F., and Zhuang, Y. (2018b).
\newblock Dynamic network embedding by modeling triadic closure process.
\newblock In {\em Proceedings of the AAAI Conference on Artificial
  Intelligence}, volume~32.

\bibitem[Zhu et~al., 2016]{zhu2016scalable}
Zhu, L., Guo, D., Yin, J., Ver~Steeg, G., and Galstyan, A. (2016).
\newblock Scalable temporal latent space inference for link prediction in
  dynamic social networks.
\newblock {\em IEEE Transactions on Knowledge and Data Engineering},
  28(10):2765--2777.

\end{thebibliography}

\end{document}